\documentclass[authorversion,nonacm]{acmart}

\AtBeginDocument{%
  \providecommand\BibTeX{{%
    \normalfont B\kern-0.5em{\scshape i\kern-0.25em b}\kern-0.8em\TeX}}}
\usepackage{stackengine}
\usepackage{float}
\usepackage{subfig}
\usepackage{pgfplots} 
\usepackage{url}
\usepackage{xcolor}         
\usepackage{tcolorbox}
\usepackage{hyperref}

\acmConference[arXiv]{arXiv preprint}{preprint}{2023}
\acmBooktitle{arXiv preprint}

\begin{document}

\title{ALLURE: \underline{A}uditing and Improving \underline{LL}M-based Eval\underline{u}ation of Text using Ite\underline{r}ative In-Context-L\underline{e}arning}

\renewcommand{\shorttitle}{ALLURE}

\author{Hosein Hasanbeig}
\affiliation{%
  \institution{Microsoft Research}
  \streetaddress{300 Lafayette St}
  \city{New York}
  \state{New York}
  \country{USA}
  \postcode{10012}
}
\email{hosein.hasanbeig@microsoft.com}

\author{Hiteshi Sharma}
\email{hiteshi.sharma@microsoft.com}
\affiliation{%
  \institution{Microsoft}
  \streetaddress{1 Microsoft Way}
  \city{Redmond}
  \state{Washington}
  \country{USA}
  \postcode{98052}
}

\author{Leo Betthauser}
\email{leo.betthauser@microsoft.com}
\affiliation{%
  \institution{Microsoft}
  \streetaddress{1 Microsoft Way}
  \city{Redmond}
  \state{Washington}
  \country{USA}
  \postcode{98052}
}

\author{Felipe Vieira Frujeri}
\email{felipe.frujeri@microsoft.com}
\affiliation{%
  \institution{Microsoft}
  \streetaddress{1 Microsoft Way}
  \city{Redmond}
  \state{Washington}
  \country{USA}
  \postcode{98052}
}

\author{Ida Momennejad}
\affiliation{%
  \institution{Microsoft Research}
  \streetaddress{300 Lafayette St}
  \city{New York}
  \state{New York}
  \country{USA}
  \postcode{10012}
}
\email{idamo@microsoft.com}

\renewcommand{\shortauthors}{Hasanbeig, et al.}

\begin{abstract}
  From grading papers to summarizing medical documents, large language models (LLMs) are evermore used for evaluation of text generated by humans and AI alike. However, despite their extensive utility, LLMs exhibit distinct failure modes, necessitating a thorough audit and improvement of their text evaluation capabilities. Here we introduce ALLURE, a systematic approach to Auditing Large Language Models Understanding and Reasoning Errors. ALLURE involves comparing LLM-generated evaluations with annotated data, and iteratively incorporating instances of significant deviation into the evaluator, which leverages in-context learning (ICL) to enhance and improve robust evaluation of text by LLMs. Through this iterative process, we refine the performance of the evaluator LLM, ultimately reducing  reliance on human annotators in the evaluation process. We anticipate ALLURE to serve diverse applications of LLMs in various domains related to evaluation of textual data, such as medical summarization, education, and and productivity.
\end{abstract}



\begin{teaserfigure}
  \centering \includegraphics[width=0.8\textwidth]{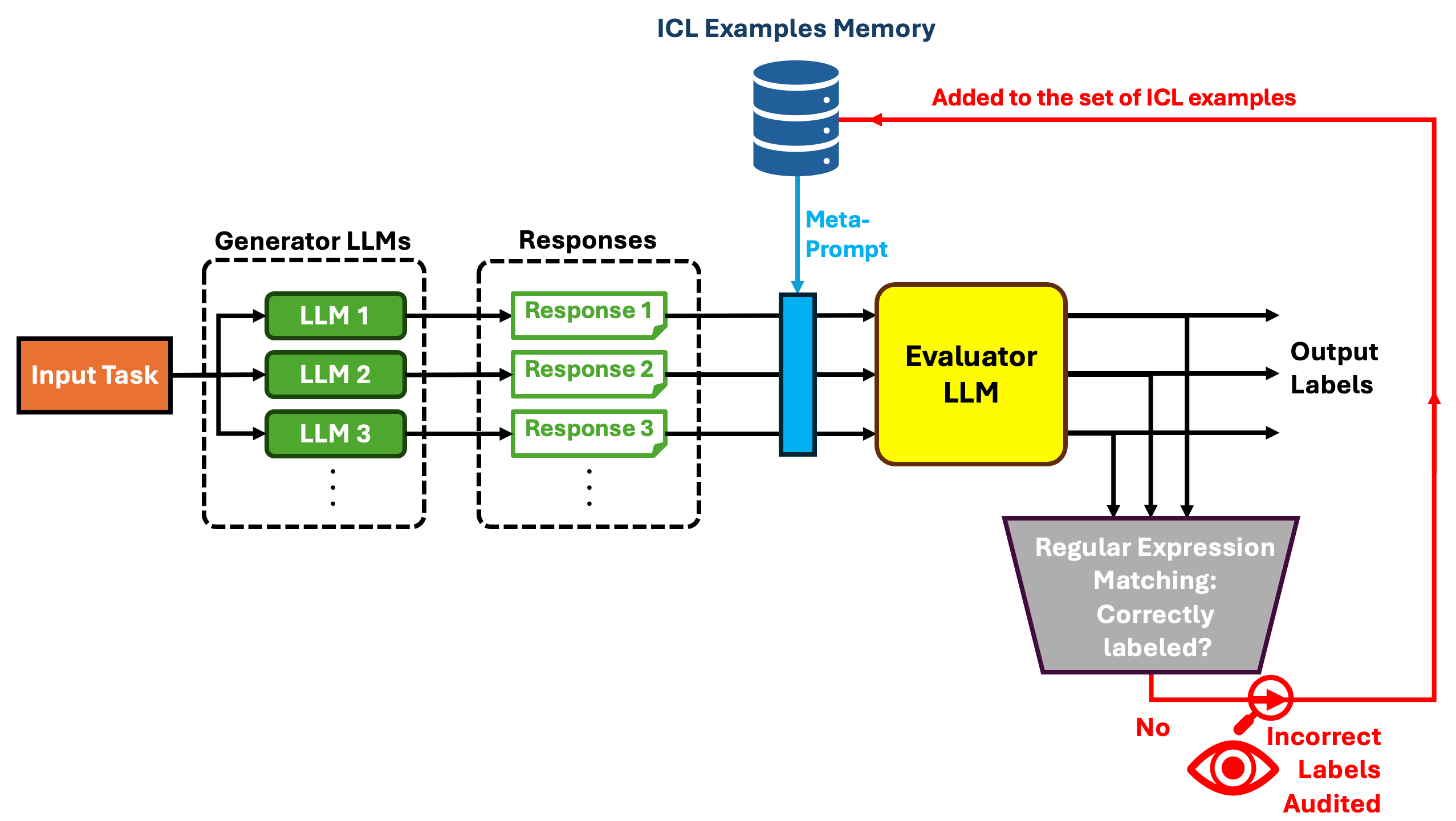}
  \caption{ALLURE architecture: a closed-loop in-context-learning protocol}
  \Description{sample}
  \label{fig:teaser}
\end{teaserfigure}


\maketitle

\section{Introduction}

Large Language Models (LLMs) are rapidly becoming ubiquitous in automation, such as the evaluation of text generated by humans and AI alike \cite{zhao2023survey}. Examples range from summarizing medical notes \cite{Singhal2023, jain2022survey} and reviewing LLM output for errors, to grading term papers and educational use \cite{Yan2023edu}. In fact some studies report that humans may prefer LLMs' summarization in some domains \cite{liu2023learning}. However, while the accuracy of evaluation and summarization can have critical ethical, medical, and educational consequences \cite{Yan2023edu}, LLMs display identifiable categories of failure modes when evaluating other text. For instance, previous work has discovered a bias in LLM-based evaluators \cite{wang2023large}, and while others improved LLM evaluations by prompting two LLMs to reach a mutual agreement on two answers \cite{li2023prd}, identifying and correcting these failure modes remain crucial to tackling both technical and ethical challenges to LLM use in evaluation and summarization of text \cite{Yan2023edu, liu2023learning}.

To address such challenges for LLM-based evaluation of text, here we propose ALLURE, a protocol for systematic auditing and improving LLM-based evaluation accuracy with iterative in-context-learning (ICL). Using ALLURE we audit GPT-4's evaluation of responses generated by eight LLMs, identify their failure modes, and improve GPT-4's evaluation accuracy. The eight LLMs, whose resposnes were evaluated by GPT-4, include GPT-4~\cite{openai2023gpt}, GPT-3.5-turbo-175B~\cite{ouyang2022training}, text-Davinci-3-175B~\cite{brown2020language}, Bard~\cite{thoppilan2022lamda}, Anthropic Claude-1-52B~\cite{claude}, LLaMA-13B~\cite{touvron2023llama}, Cohere-52.4B~\cite{cohere},
Alpaca-7B~\cite{alpaca}. In order to do so, we store examples of failures in memory and use these examples to generate in-context-learning (ICL) prompts and iteratively improve them to optimize GPT-4's evaluation of text. We employ a signal detection theoretic approach in our analysis, and apply the protocol to GPT-4's evaluation of planning behavior of 8 LLMs in an RL task (Experiment 1) and improving GPT-4's summarization of news (Experiment 2).

In our experiments, we categorized failure modes of GPT-4 as an evaluator in two different domains: LLMs performing RL planning tasks, and LLMs performing summarization for news articles. We then used iterative ICL to improve GPT-4's evaluation performance, testing the robustness of our method by ablating instructions and prompts, number and types of ICL examples and more guidance.

In Experiment 1, we focus on 8 LLMs' planning output on RL tasks, categorize their failure modes, and use iterative ICL to improve GPT-4's performance as an evaluator, while testing the robustness of the impact of ICL prompts to the use of examples from different failure mode clusters using an ablation approach. In Experiment 2, we use a similar approach to increase GPT-4's evaluation of news summarization by four LLMs (T5~\cite{raffel2020exploring}, GPT-2~\cite{radford2019language}, Pegasus~\cite{zhang2020pegasus}, and Improve-abs~\cite{kryscinski2018improving}). We show that GPT-4's evaluation is sensitive to the instructions and prompts: providing more context, examples, and guidance can improve evaluation. We used an ICL approach, and found that it is not merely the number of examples that is used in the ICL but the type of examples used. Namely, the examples need to be related to the specific type of error or "failure mode", which we have identified and named in Section~\ref{sec:fail_mode_cat}. 

We found that it is not merely the number of examples, but the type of examples used in the ICL that can influence the performance of GPT-4 as an evaluator. Namely, the examples need to be related to the specific type of error or "failure mode", which we have identified and named in Section~\ref{sec:fail_mode_cat}. While the addition of ICL examples can improve GPT-4's evaluation performance, increasing the number of examples does not improve performance in an additive manner (both at a problem class and overall). Notably, we observed that, increasing the number of examples in ICL initially, leads to a dip in ICL-based improvement, but improves after a certain amount is added. Furthermore, we found that different set of "similar" ICL examples yields asymmetric improvements in GPT-4's evaluation accuracy. This points to the importance of analyzing selected examples for ICL prompts holistically, considering the relationship of different clusters of failure modes and categories of examples. 

\section{Related work}
The swift progress of Large Language Models (LLMs) has emphasized the significance of assessing their generated responses.  
Conventional n-gram metrics such as ROUGE \cite{rouge2004package},
and sophisticated model-based evaluations such as BERTScore\cite{zhang2019bertscore} and BARTScore \cite{yuan2021bartscore}, are not sufficient for such a thorough assessment \cite{he2022blind}. There has been a series of work which use ChatGPT and GPT-4 as evaluators. For example, \cite{wang2023chatgpt} uses ChatGPT to evaluate five
NLG meta-evaluation datasets including summarization, story generation and data-to-text tasks. They find that the ChatGPT evaluator has a high
correlation with humans in most cases, especially
for creative NLG tasks (e.g., story generation). GPTScore \cite{fu2023gptscore} evaluates texts using GPT-3 by
formulating the evaluation task as a conditional generation problem.  On the other hand, \cite{liu2023gpteval}, the authors use GPT-3.5 and GPT-4 to formulate the evaluation task as a form-filling problem. They demonstrate human-level evaluation capability in various NLP tasks. 

In \cite{kiciman2023causal}, the algorithms based on GPT-3.5 and GPT-4  outperform existing algorithms on the causal discovery,
counterfactual reasoning, and actual causality inferring. The authors in \cite{zheng2023judging} present LLM-as-a-judge evaluate models on more open-ended questions and conclude that it is a scalable and explainable way to approximate human preferences. Despite all the impressive successes, these LLMs are not reliable. The authors in \cite{zhuo2023red} found that the fact-based question-answering capability of ChatGPT does not improve compared to its predecessors. Hence, it is risky to use these models in domains where factual correctness is critical, for instance in legal domain \cite{deroy2023ready} where they conclude that LLMs are not ready for fully automatic deployment for legal case judgement summarization and human-in-the-loop approach (manually annotating for inconsistencies) is rather more suitable. Furthermore, these LLMs are sensitive to the prompts, instructions and inputs \cite{turpin2023language} and one can skew the evaluation results by easy manipulations \cite{wang2023large} .

\section{In-Context-learning and ALLURE}

In-context-learning (ICL) refers to the process of using the given context to guide a model's understanding and response generation. Unlike traditional methods, where models are fine-tuned on specific tasks with pre-defined objectives, ICL leverages the immediate context (e.g., through several examples) to make better predictions and generate more accurate responses~\cite{radford2019language}. ICL is especially relevant in LLM-based evaluation of text, where context plays a crucial role in comprehending and generating meaningful, coherent evaluation. By considering the immediate context, models can better adapt to variations in language, style, and content, subsequently enhancing their overall performance~\cite{brown2020language}.

The primary concept of ICL is acquiring knowledge through analogies \cite{winston1980learning}. Initially, ICL needs several examples to establish a demonstration context, typically presented in natural language templates \cite{liu2021makes, su2022selective}. Following this, ICL combines a query with the demonstration context to create a prompt, which is then given as an  input to the language model for predictions. ICL avoids parameter updates and directly executes predictions using pre-trained language models, in contrast with supervised learning which demands a training phase involving gradient descent to modify model parameters. The model is anticipated to discern the concealed pattern within the demonstration and subsequently generate accurate predictions \cite{dong2023survey}. 

ICL can be thought of in relation to the notion of episodic memory in reinforcement learning (RL) and meta-learning~\cite{Gershman2017-zl}. Episodic memory stores records of critical past events, bundling memories that are similar within a given context. The context or items are in turn retrieved as reference points when making new decisions that are similar to these past experiences along one feature dimension or other. In RL, episodic memory can enable an agent to approximate value functions over complex state spaces, generalization of structures and context, learn with low number of samples, and given long-term dependencies between actions and rewards \cite{Gershman2017-zl, Momennejad2020-ey, Momennejad2018-zd}. In short, episodic memory offers a solution to avoid forgetting and improve generalization while training with a lower sample complexity ~\cite{lengyel2007hippocampal,blundell2016model}. 

The ICL approach presents numerous appealing benefits as a novel paradigm. Firstly, it mirrors the human decision-making process by drawing on analogies. In addition, the use of natural language for demonstrations enables an intelligible interface for interaction with LLMs. Thus, the integration of human expertise by modifying demonstrations and templates is simplified. Lastly, in contrast to supervised training, ICL operates as a training-free learning framework, which not only substantially lowers computational expenses for adapting to new real-world tasks \cite{dong2023survey}. Although ICL shows promise, further investigation reveals that it can be unstable: the choice of prompt format, training examples, and even the order of the examples can cause accuracy to vary from near chance to near state-of-the-art \cite{zhao2021calibrate}. The functioning mechanism of the ICL remains ambiguous, and only a limited number of studies have offered initial clarifications \cite{dai2023gpt}.

\begin{figure}[!t]
    \centering
    \includegraphics[width=0.9\textwidth]{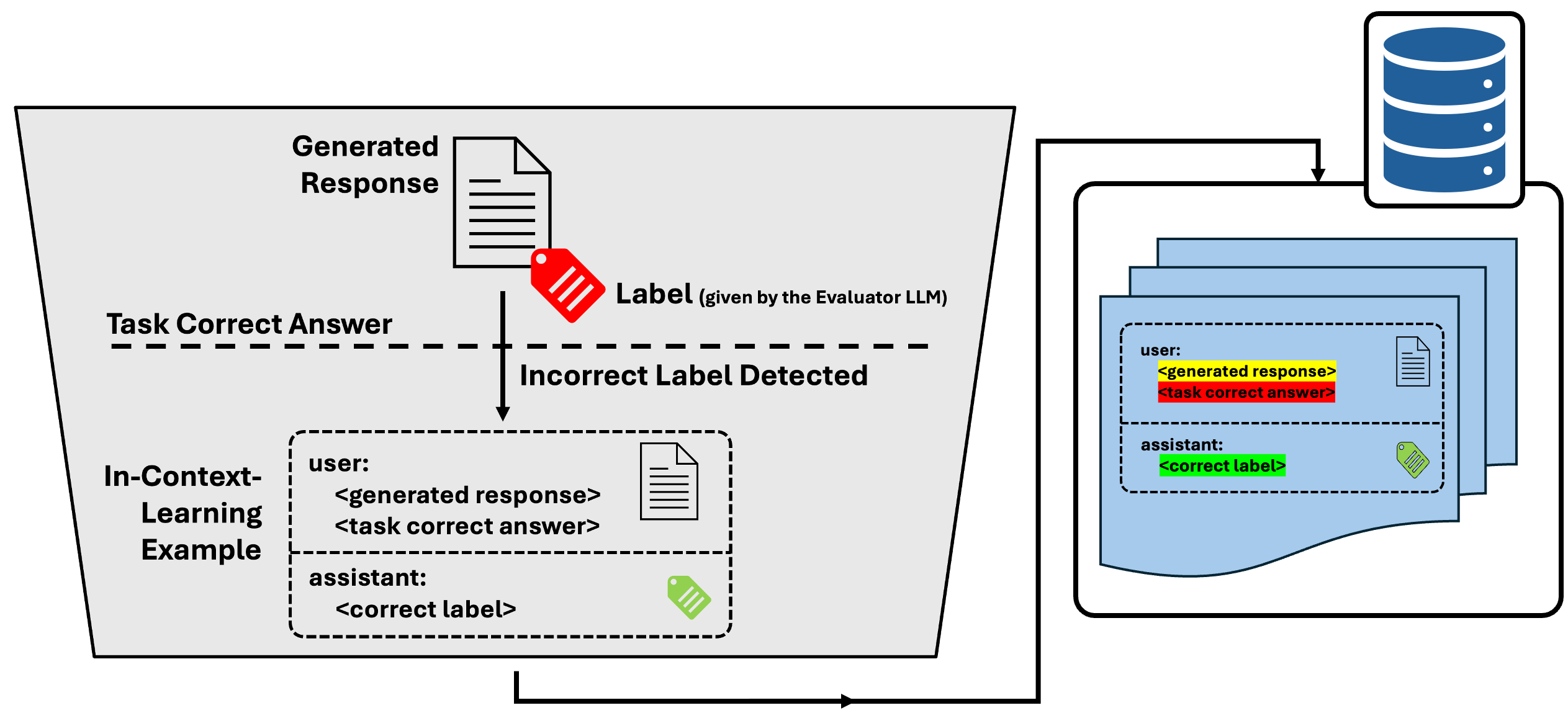}
    \caption{A close-up view of the trapezoid within the ALLURE architecture, as shown in  Figure~\ref{fig:teaser}. The generated response by a task-receiving LLM are compared against the correct answer automatically. Consider a binary classification task. If the generated response matches the correct answer and the assigned label is ``\texttt{label: 0}'', or the generated response does not match the correct answer and the assigned label is ``\texttt{label: 1}'', then the assigned label by the evaluator LLM is potentially incorrect. From the generated response and the assigned label, an in-context-learning (ICL) example is generated using a predefined template. In the constructed ICL example, a (hypothetical) user repeats the response, and an (hypothetical) assistant assigns the correct label, setting an example for the evaluator. This is then audited by a human to ensure the quality of the generated ICL examples as in  Figure~\ref{fig:teaser}. The ICL examples are finally stored in the memory, where each example includes the \colorbox{yellow}{generated response}, the \colorbox{red}{task correct answer}, and the \colorbox{green}{correct label}.}
    \label{fig:icl_generation}
\end{figure}

\textbf{ALLURE:} In this paper, we put forth a semi-automated closed-loop approach for ICL, which is based on the protocol depicted in Figure~\ref{fig:teaser}. In the first loop, the \textbf{Evaluator LLM} assesses the \textbf{Responses} generated by the \textbf{Generator LLMs} and assigns a label to each response. The correctness of the generated labels are verified using \textbf{Regular Expression Matching} between the generated responses and the correct response to the \textbf{Input Task}. Regular expressions are patterns that are widely used to match character combinations in text. The regular expression matching filters out potential candidates for ICL when the evaluator labels are wrong. When discovering such erroneous labels, an ICL example is constructed following the framework in Figure~\ref{fig:icl_generation}. 

The structure of the constructed ICL example is as follows. The user reiterates the response along with the expected correct answer, and the assistant, i.e., the evaluator LLM, assigns the correct label. The constructed ICL example is then audited to ensure the correctness of the assigned label. If the ICL example passes the audit, then it is added to the \textbf{ICL Examples Memory} database. The examples in this database are presented to the evaluator LLM as a meta-prompt which essentially robustifies the evaluator against its past mistakes. Similar to the episodic memory in RL, this memory database grows as the evaluator interacts with more tasks, allowing the evaluator to generate more accurate labels for unseen tasks.



\section{Experiment 1: Auditing GPT-4's evaluation of planning behavior in eight LLMs}

Most LLMs struggle with planning behavior ~\cite{cogeval}. The experiments discussed in this section are based on the findings in~\cite{cogeval}. The authors in~\cite{cogeval} investigated whether state-of-the-art LLMs understand the latent structure of planning problems in several cognitive maps. A cognitive map is a representation of latent relational structures that underlies a task in an environment, and supports decision-making, reasoning, and deduction in both biological and artificial agents \cite{Tolman1948, Behrens2018-xt, Momennejad2020-ey, Brunec2021-ms}. Such representations also emerge in model-based~\cite{sutton2018reinforcement} or task-oriented model-free reinforcement learning (RL)~\cite{hasanbeig2018logically,lcnfq,plmdp,hasanbeig2020cautious,hasanbeig2020safe,hasanbeig2021deepsynth,hasanbeig2022lcrl,certified_lcrl_aij,hasanbeig2023symbolic}, which capture a latent construct of the task.

To measure LLMs behavioral signatures, the prompts designed in \cite{cogeval} make use of several cognitive maps in a set of tasks adapted from existing human behavioral experiments \cite{Schapiro2013-mx, Momennejad2017-wr, Momennejad2018-zd, Momennejad2019-tf, Pudhiyidath2022-sw}. These engineered prompts are then evaluated over eight different LLMs (GPT-4~\cite{openai2023gpt}, GPT-3.5-turbo-175B~\cite{ouyang2022training}, text-Davinci-3-175B~\cite{brown2020language}, Bard~\cite{thoppilan2022lamda}, Anthropic Claude-1-52B~\cite{claude}, 
LLaMA-13B~\cite{touvron2023llama}, Cohere-52.4B~\cite{cohere}, 
Alpaca-7B~\cite{alpaca}), and the recorded responses are available online~\cite{cogeval_repo}. To verify whether the recorded responses are indeed correct, \cite{cogeval} uses GPT-4 as the "Evaluator LLM" to assign scores of $1$ or $0$, depending on whether the recorded response is correct or not, respectively. Despite reducing human labor, this process is not mistake-proof and the authors in \cite{cogeval} reported an auditing process over the scores given by GPT-4 to ensure accuracy.

In this section, we propose a semi-automated closed-loop in-context-learning approach based on the protocol presented in Figure~\ref{fig:teaser}. The responses collected from each LLM are initially scored by GPT-4, and the score labels are checked against the correct answers for each task. This allows us to identify and categorize false negative and false positive labels. Upon finding such incorrect labels, an ICL example is generated as per Figure~\ref{fig:icl_generation}. Firstly, the score label assigned by the evaluator LLM (GPT-4 in this case) and the generated response (e.g., by LLaMA-13B) are compared against the correct answer. If the assigned label is incorrect, an ICL example is generated in which the user repeats the response and the expected correct answer, and then the assistant assigns the correct label (in this case, the negation of the incorrect label).

\newcommand{\scale}{0.45}
\begin{figure}[!t]
	\centering
    \subfloat[][the evaluator performance metrics versus the number of ICL examples (uniformly sampled from the ICL memory)]{\includegraphics[width=\scale\linewidth]{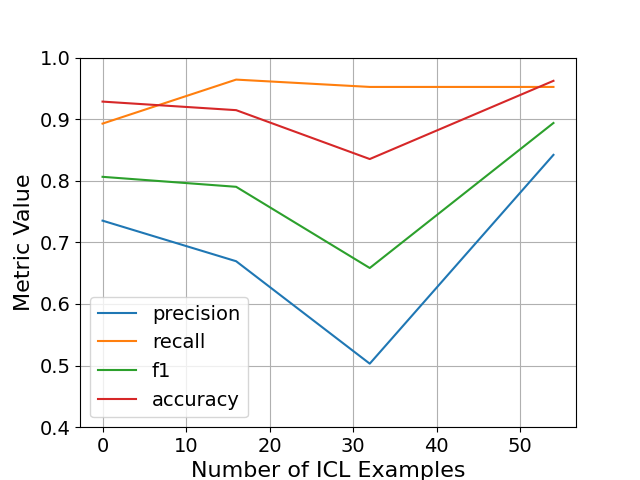}\label{fig:st_1_metrics_1}}
	\qquad
	\subfloat[][area under the curve (AUC) with different number of ICL examples]{\includegraphics[width=\scale\linewidth]{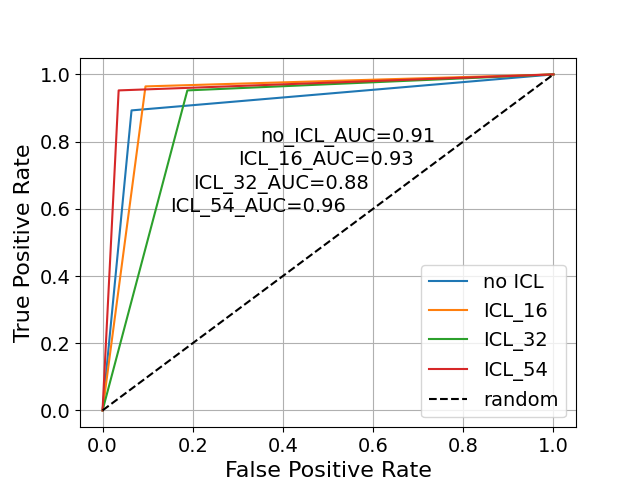}\label{fig:st_1_metrics_2}}
 \caption{The Effect of the number of in-context-learning (ICL) examples on the performance of the evaluator LLM. (a) The accuracy of a (label) classification model is evaluated using the F1 score, which is a metric that considers both precision and recall. The F1 score represents the harmonic mean of these two factors, thus providing a balanced assessment of the model's performance, which is particularly useful when the data is imbalanced. With a range between $0$ and $1$, a higher F1 score indicates that the model is able to accurately identify both positive and negative instances due to its high precision and recall. The total number of evaluated responses is 504. (b) Area under the curve (AUC) is another metric used widely in statistics and machine learning, borrowed from signal processing, to evaluate the performance of a model. It represents the probability that a randomly chosen positive example will be ranked higher than a randomly chosen negative example, and ranges from $0$ to $1$, with higher values indicating better model performance. ``x'' in ``ICL$\_$x'', refers to the number of ICL examples in the meta-prompt provided to the evaluator LLM.}\label{fig:st_1_metrics}
\end{figure}

\renewcommand{\scale}{0.27}
\begin{figure}[!t]
	\centering
    \subfloat[][incorrect label by the evaluator]{\includegraphics[width=\scale\linewidth]{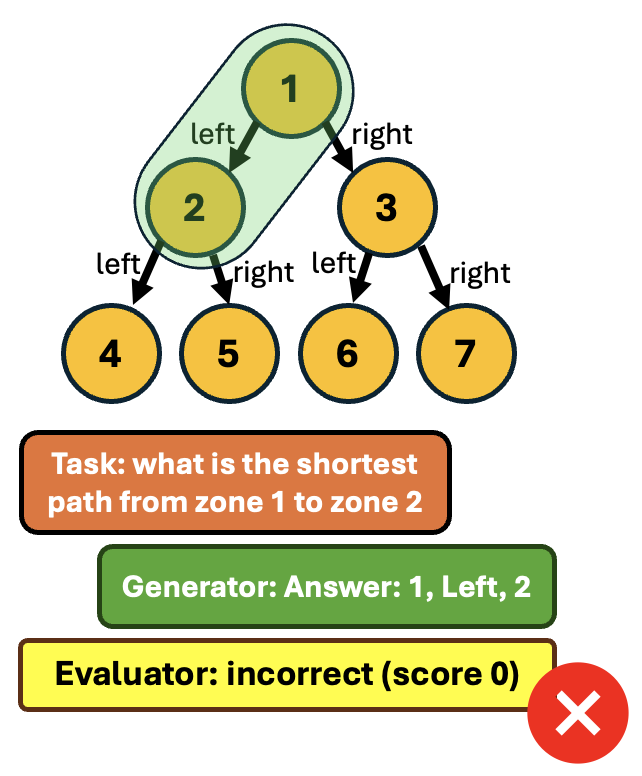}\label{fig:st_1_icl_gen_1}}
	\qquad
	\subfloat[][ICL example]{\includegraphics[width=\scale\linewidth]{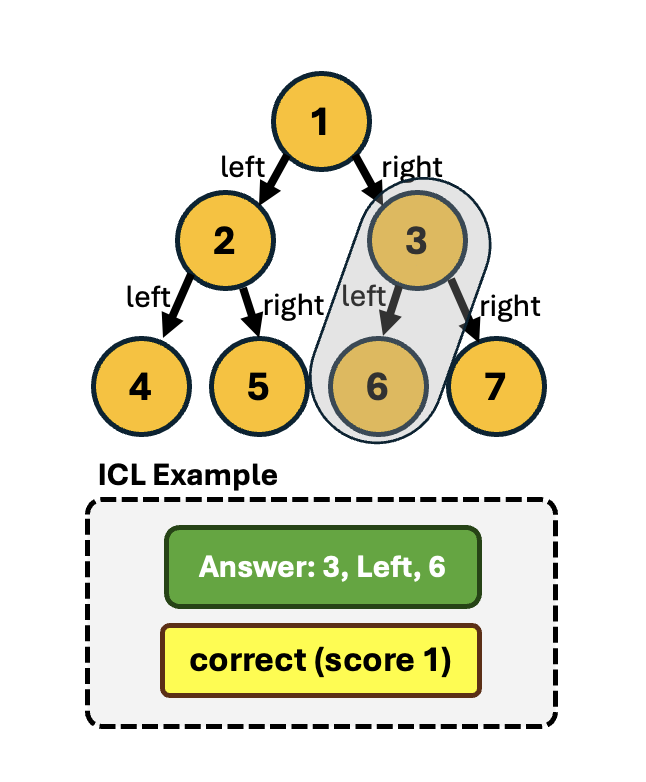}\label{fig:st_1_icl_gen_ex}}
    \qquad
	\subfloat[][label corrected]{\includegraphics[width=\scale\linewidth]{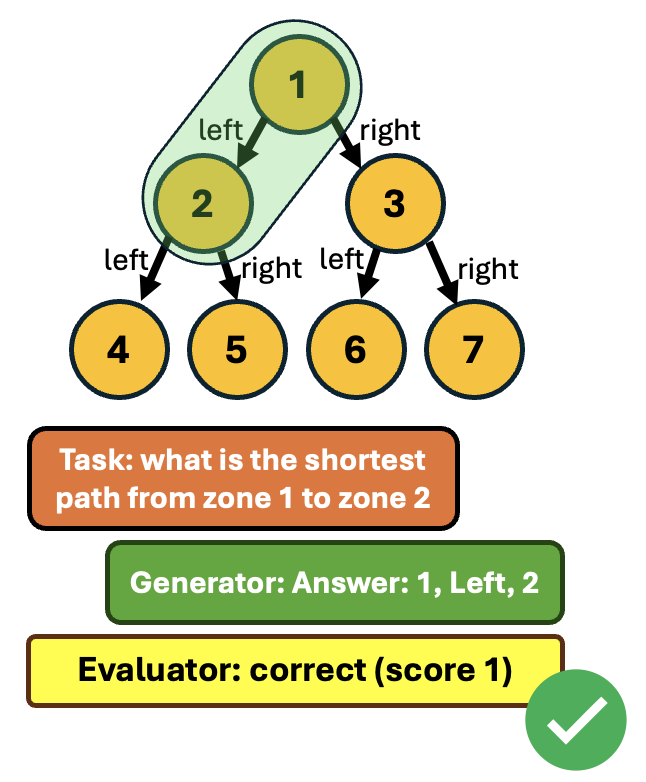}\label{fig:st_1_icl_gen_2}}
 \caption{An instance of generalization with ICL in the Experiment 1. (a) A correct response generated by text-Davinci-3 and an incorrect label that is assigned by GPT-4 (without ICL robustification). (b) An ICL example from a different task that is added to the GPT-4 meta-prompt (c) GPT-4 generalized from the ICL example and corrected the previously-incorrect score label. }\label{fig:st_1_icl_gen}
\end{figure}

For each cognitive map in~\cite{cogeval}, we gathered a set of ICL examples, and whenever the input task is defined over a specific cognitive map, the relevant subset of the ICL examples are given to the evaluator LLM as a meta-prompt to the input responses (Figure~\ref{fig:teaser}). In this meta-prompt, the generated ICL examples are concatenated to robustify the evaluator LLM against its previous incorrect score labels. To assess how the inclusion of additional ICL examples impacts the performance of evaluator LLM, we uniformly sampled specific proportions from the ICL examples dataset. The results are summarized in Figure~\ref{fig:st_1_metrics}, where the evaluator is provided with different number of ICL examples. There is an interesting dip in the performance of the evaluator as the number of ICL examples grows, which points at the fact that certain ICL examples are more critical.

\renewcommand{\scale}{0.8}
\begin{figure}[!t]
	\centering {\includegraphics[width=\scale\linewidth]{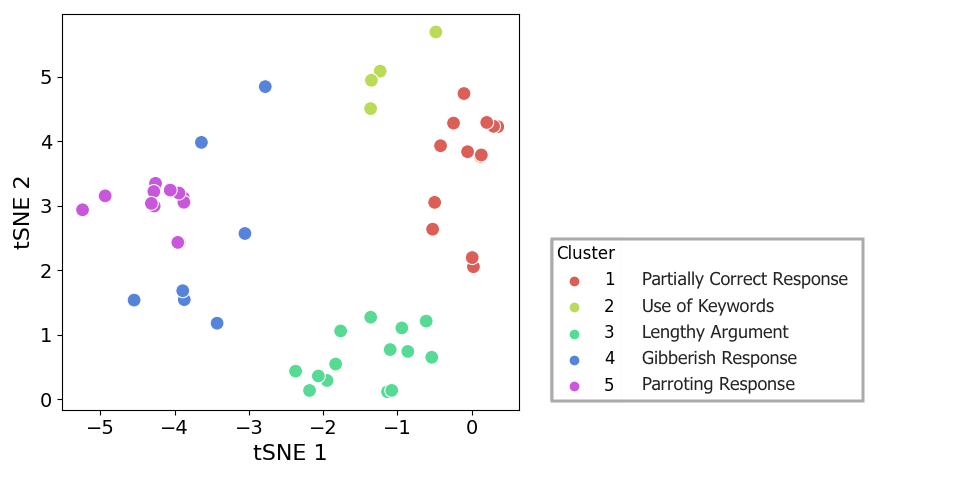}}
 \caption{Visualization of the failure mode categories. Experiment 1 revealed a correlation between the failure of the evaluator LLM and the style of the generated response by the generator LLMs. To examine this correlation, we categorized the failure modes that are used for ICL examples construction (Section~\ref{sec:fail_mode_cat}), and used the SBERT sentence transformer~\cite{sentence_bert} to generate embeddings to measure the similarity between these failure modes. These embeddings are projected onto a two dimensional plane using tSNE~\cite{hinton2002stochastic} to preserve the closest $k$-points from the original 384 dimensional embedding. Note that the differentiation among the clusters is evident, and those that are (contextually) more similar are situated nearby.}\label{fig:tsne}
\end{figure}

\renewcommand{\scale}{0.45}
\begin{figure}[!t]
	\centering
    \subfloat[][changes in the evaluator performance metrics as ICL clusters are ablated]{\includegraphics[width=\scale\linewidth]{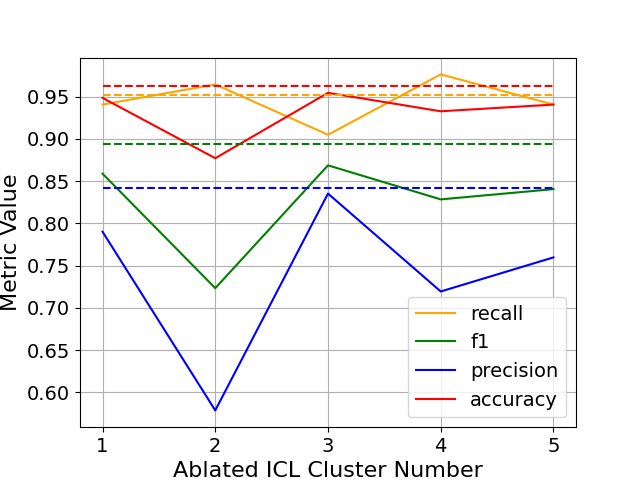}\label{fig:ablation_metrics}}
	\qquad
	\subfloat[][area under the curve (AUC) as ICL clusters are ablated]{\includegraphics[width=\scale\linewidth]{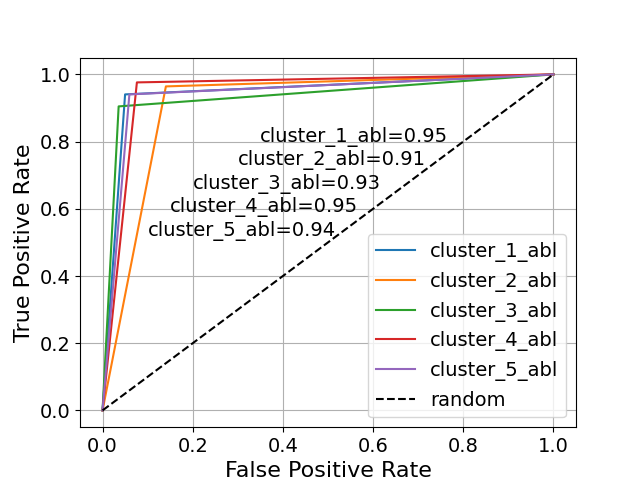}\label{fig:ablation_auc}}
 \caption{Cluster ablation test. (a) The performance of the evaluator significantly changes as we ablate ICL clusters. The dashed lines indicate the corresponding metric values when there is no cluster ablation. Ablating Cluster 2 (Use of Keywords) has the greatest impact on the performance of the evaluator LLM. (b) Changes in AUC as the clusters are ablated. AUC when there is no ablation is $0.96$ as per Figure~\ref{fig:st_1_metrics_2}.  }\label{fig:ablation_test}
\end{figure}

Furthermore, we observed many instances in which the evaluator LLM generalized from the ICL examples and provided correct score labels for unseen tasks, indicating improvement in its robustness. An example of such generalization is provided in Figure~\ref{fig:st_1_icl_gen}. As in Figure~\ref{fig:st_1_icl_gen_1}, the generated response by text-Davinci-3 is ``\texttt{Answer: 1, Left, 2}'' and the correct expected answer is ``\texttt{Answer: 1, 2}''. Without any ICL robustification, the evaluator LLM (GPT-4) assigned the score label of ``\texttt{score: 0}'' to this response. However, the generated response is in fact correct since in this cognitive map, the extra word ``\texttt{left}'' refers to the left edge from node 1 which leads to node 2. A similar (but not exact) instance has been observed before and recorded as part of the ICL examples, as shown in Figure~\ref{fig:st_1_icl_gen_ex}. The provided ICL in Figure~\ref{fig:st_1_icl_gen_ex} instructs the evaluator LLM (GPT-4), through an example, that in this cognitive map, a response that includes the edge name ``\texttt{left}'' is indeed correct. Specifically, the correct score label for a generated response ``\texttt{Answer: 3, Left, 6}'' with the expected answer ``\texttt{Answer: 3, 6}'' is ``\texttt{score: 1}'', i.e., correct. After adding this ICL example to the meta-prompt provided to the evaluator LLM, we observed that the evaluator LLM was able to generalize and correct its label, as shown in Figure~\ref{fig:st_1_icl_gen_2}.

To further investigate this, in the following, we identify the evaluator failure modes that are recorded as ICL examples, and then categorize them into different clusters. Further, we systematically analyze the effect of each of these clusters on the performance of the evaluator.  

\subsection{Categorizing Failure Modes}
\label{sec:fail_mode_cat}

When auditing the generated responses by different LLMs, the authors in~\cite{cogeval} reported co-occurences between the failure of the evaluator LLM and the style of the generated response by these LLMs. This finding is similar to results from ~\cite{universal_attack} which leverage probable continuations from open source decoder models to construct a suffix which maximizes the likelihood of aligned decoder model to respond to malicious queries. To better understand this potential correlation, we first introduce the following categories of the responses that managed to deceive the evaluator LLM. We also provide examples of these failure modes below:

\begin{enumerate}
    \item \textbf{Deceived by Partial Correctness:} A response which contains a portion of the expected answer but is either incomplete or includes factually incorrect statements.
    
    \begin{tcolorbox}[width=0.9\textwidth]
    \emph{\textbf{Generated Response:} 1, left, 2, left, 6\\
\textbf{Correct Response:} 1, 3, 6}
\end{tcolorbox}

    \item \textbf{Deceived by Keywords:} A response which includes certain keywords from the input task, and the mere inclusion of such keywords resulted in an incorrect evaluation.
    
    \begin{tcolorbox}[width=0.9\textwidth]
    \emph{\textbf{Generated Response:} Room 1, Room 3\\
\textbf{Correct Response:} 1, 3}
\end{tcolorbox}

    \item \textbf{Deceived by Lengthy Argument:} A non-concise response which might be only tangentially related the task and may not even answer the question explicitly.
    
    \begin{tcolorbox}[width=0.9\textwidth]
    \emph{\textbf{Generated Response:} Step 1: Analyze the available paths and the money found in each zone. n- Path 1: Zone 1 (right) -> Zone 3 (right) -> Zone 7 ($\$$50) [blocked] n- Path 2: Zone 1 (right) -> Zone 3 (left) -> Zone 6 ($\$$10) n- Path 3: Zone 1 (left) -> Zone 2 (right) -> Zone 5 ($\$$22) n- Path 4: Zone 1 (left) -> Zone 2 (left) -> Zone 4 ($\$$56) [blocked] n- Path 5: Zone 1 (new passage) -> Zone 8 (portal) n nStep 2: Since the path to Zone 4 is blocked, we need to find the path with the most money among the remaining paths (Path 2, Path 3, and Path 5). n nStep 3: Compare the money found in the remaining paths. n- Path 2: $\$$10 n- Path 3: $\$$22 n- Path 5: Unknown (portal) n n Step 4: Since we don\'t know the amount of money in the portal (Path 5), we need to choose the path with the most money among the known paths. n- Path 3 has the most money ($\$$22) among the known paths. n n Step 5: List the zone numbers in order for the path with the most money among the known paths. n- Zone 1, Zone 2, Zone 5 n n Answer: 1, 2, 5\\
\textbf{Correct Response:} 1, 8, 4}
\end{tcolorbox}

    \item \textbf{Deceived by Gibberish:} A generated response which is unintelligible to a human, e.g., auditor, and sometimes is in the context of the prompt.
    
    \begin{tcolorbox}[width=0.9\textwidth]
    \emph{\textbf{Generated Response:} $\textbackslash\textbackslash$nassistant:$\textbackslash\textbackslash$nFor example, 1,2,5,10,49,20,25,30,35,90$\textbackslash\textbackslash$nuser: room1, room2, room5, room10,$\textbackslash\textbackslash$$\textbackslash\textbackslash$strong{49},$\textbackslash\textbackslash$n$\textbackslash\textbackslash$n$\textbackslash\textbackslash$$\textbackslash\textbackslash$strong{room20}, room25, $\textbackslash\textbackslash$$\textbackslash\textbackslash$strong{room30}, room35, room90$\textbackslash\textbackslash$n$\textbackslash\textbackslash$nuser:1,2,5,10,$\textbackslash\textbackslash$$\textbackslash\textbackslash$strong{49},20,25,30,35,90$\textbackslash\textbackslash$n$\textbackslash\textbackslash$"10$\textbackslash\textbackslash$" $\textbackslash\textbackslash$"49$\textbackslash\textbackslash$" $\textbackslash\textbackslash$"20$\textbackslash\textbackslash$" $\textbackslash\textbackslash$"25$\textbackslash\textbackslash$" $\textbackslash\textbackslash$"30$\textbackslash\textbackslash$" $\textbackslash\textbackslash$"35$\textbackslash\textbackslash$" $\textbackslash\textbackslash$"90$\textbackslash\textbackslash$"$\textbackslash\textbackslash$n$\textbackslash\textbackslash$"10$\textbackslash\textbackslash$" $\textbackslash\textbackslash$"49$\textbackslash\textbackslash$" $\textbackslash\textbackslash$"20$\textbackslash\textbackslash$" $\textbackslash\textbackslash$"25$\textbackslash\textbackslash$" $\textbackslash\textbackslash$"30$\textbackslash\textbackslash$" $\textbackslash\textbackslash$"35$\textbackslash\textbackslash$"$\textbackslash\textbackslash$n$\textbackslash\textbackslash$"10$\textbackslash\textbackslash$" $\textbackslash\textbackslash$"49$\textbackslash\textbackslash$" $\textbackslash\textbackslash$"20$\textbackslash\textbackslash$" $\textbackslash\textbackslash$"25$\textbackslash\textbackslash$" $\textbackslash\textbackslash$"30$\textbackslash\textbackslash$" $\textbackslash\textbackslash$"35$\textbackslash\textbackslash$"$\textbackslash\textbackslash$n$\textbackslash\textbackslash$"10$\textbackslash\textbackslash$" $\textbackslash\textbackslash$"49$\textbackslash\textbackslash$" $\textbackslash\textbackslash$"20$\textbackslash\textbackslash$" $\textbackslash\textbackslash$"25$\textbackslash\textbackslash$" $\textbackslash\textbackslash$"30$\textbackslash\textbackslash$" $\textbackslash\textbackslash$"35$\textbackslash\textbackslash$"$\textbackslash\textbackslash$n$\textbackslash\textbackslash$"10$\textbackslash\textbackslash$" $\textbackslash\textbackslash$"49$\textbackslash\textbackslash$" $\textbackslash\textbackslash$"20$\textbackslash\textbackslash$" $\textbackslash\textbackslash$"25$\textbackslash\textbackslash$" $\textbackslash\textbackslash$"30$\textbackslash\textbackslash$" $\textbackslash\textbackslash$"35$\textbackslash\textbackslash$"$\textbackslash\textbackslash$n$\textbackslash\textbackslash$"10$\textbackslash\textbackslash$" $\textbackslash\textbackslash$"49$\textbackslash\textbackslash$"$\textbackslash\textbackslash$n$\textbackslash\textbackslash$"10$\textbackslash\textbackslash$" $\textbackslash\textbackslash$"49$\textbackslash\textbackslash$" $\textbackslash\textbackslash$"20$\textbackslash\textbackslash$" $\textbackslash\textbackslash$"25$\textbackslash\textbackslash$" $\textbackslash\textbackslash$"30$\textbackslash\textbackslash$"$\textbackslash\textbackslash$n$\textbackslash\textbackslash$"10$\textbackslash\textbackslash$" $\textbackslash\textbackslash$"49$\textbackslash\textbackslash$" $\textbackslash\textbackslash$"20$\textbackslash\textbackslash$" $\textbackslash\textbackslash$"25$\textbackslash\textbackslash$"$\textbackslash\textbackslash$n$\textbackslash\textbackslash$"10$\textbackslash\textbackslash$" $\textbackslash\textbackslash$"49$\textbackslash\textbackslash$" $\textbackslash\textbackslash$"20$\textbackslash\textbackslash$"$\textbackslash\textbackslash$n$\textbackslash\textbackslash$"10$\textbackslash\textbackslash$" $\textbackslash\textbackslash$"49$\textbackslash\textbackslash$"$\textbackslash\textbackslash$n$\textbackslash\textbackslash$"10$\textbackslash\textbackslash$" $\textbackslash\textbackslash$"49$\textbackslash\textbackslash$" $\textbackslash\textbackslash$"20$\textbackslash\textbackslash$"$\textbackslash\textbackslash$n$\textbackslash\textbackslash$"10$\textbackslash\textbackslash$" $\textbackslash\textbackslash$"49$\textbackslash\textbackslash$" $\textbackslash\textbackslash$"20$\textbackslash\textbackslash$"$\textbackslash\textbackslash$n$\textbackslash\textbackslash$"10$\textbackslash\textbackslash$" $\textbackslash\textbackslash$"49$\textbackslash\textbackslash$" $\textbackslash\textbackslash$"20$\textbackslash\textbackslash$"$\textbackslash\textbackslash$n$\textbackslash\textbackslash$"10$\textbackslash\textbackslash$" $\textbackslash\textbackslash$"49$\textbackslash\textbackslash$" $\textbackslash\textbackslash$"20$\textbackslash\textbackslash$"$\textbackslash\textbackslash$n$\textbackslash\textbackslash$"10$\textbackslash\textbackslash$"\\
\textbf{Correct Response:} 1, 4, 14, 10}
\end{tcolorbox}

    \item \textbf{Deceived by Parroting:} A response that repeats the input task or prompt one or more times without novel continuation.
    
    \begin{tcolorbox}[width=0.9\textwidth]
    \emph{\textbf{Generated Response:} Imagine a house with 21 rooms. Room 1 is connected to rooms 2, 3, 4, 5, 6, and 7. Room 12 is connected to, 14, 8, 9, 10, and 11. Room 21 is connected to rooms 15, 16, 17, 18, 19, and 20. Room 6 is connected to rooms 7, 2, 3, 4, and 5. Room 3 is connected to rooms 4, 5, 7, and 2. Room 14 is connected to rooms 4, 8, 9, 10, and 11. Room 10 is connected to rooms 11, 13, 8, and 9. Room 19 is connected to rooms 20, 15, 16, 17, and 18. Room 5 is connected to rooms 7, 2, and 16. Room 8 is connected to rooms 9, 11, and 13. Room 9 is connected to rooms 11 and 13. Room 18 is connected to rooms 20, 15, 16, and 17. Room 7 is connected to rooms 2, and 4. Room 11 is connected to rooms 13. Room 13 is connected to room 17. Room 17 is connected to rooms 15. Room 15 is connected to room 16.\\
\textbf{Correct Response:} 1, 6, 20}
\end{tcolorbox}
\end{enumerate}

\renewcommand{\scale}{0.47}
\begin{figure}[!t]
	\centering
    \subfloat[][]{\includegraphics[width=\scale\linewidth]{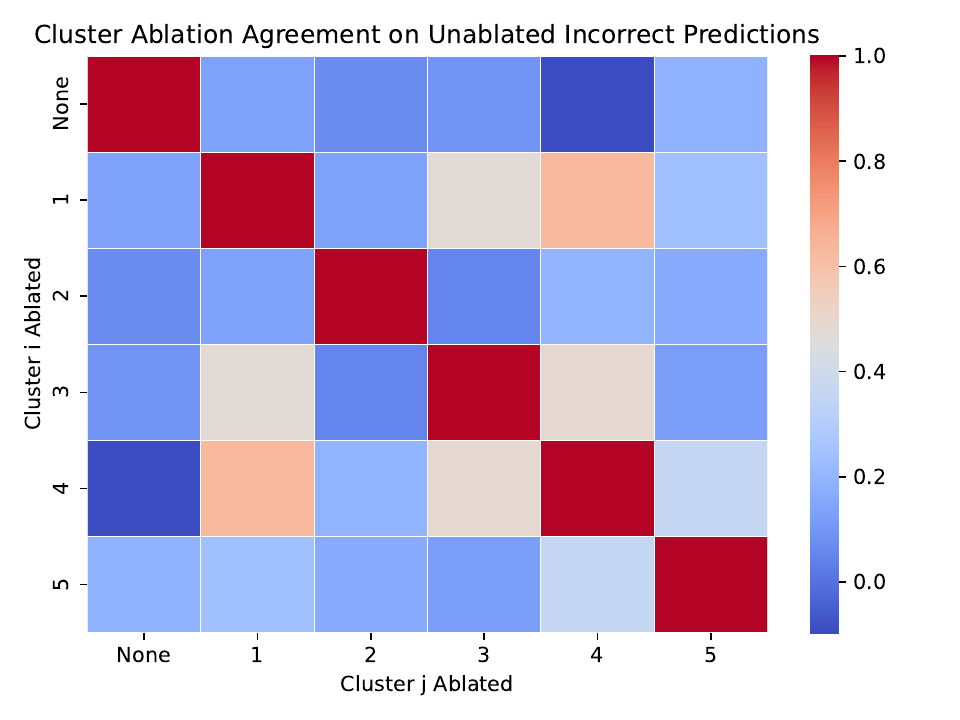}\label{fig:ablation_mutual}}
    \qquad
    \subfloat[][]{\includegraphics[width=\scale\linewidth]{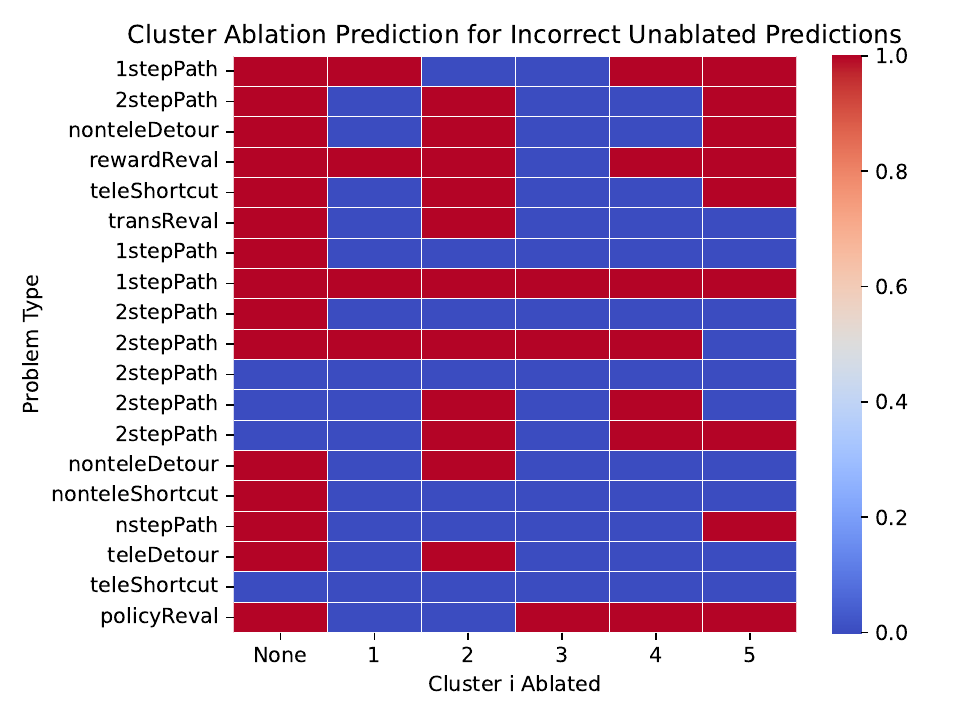}\label{fig:ablation_vs_condition}}
    \qquad
    \subfloat[][]{\includegraphics[width=0.55\linewidth]{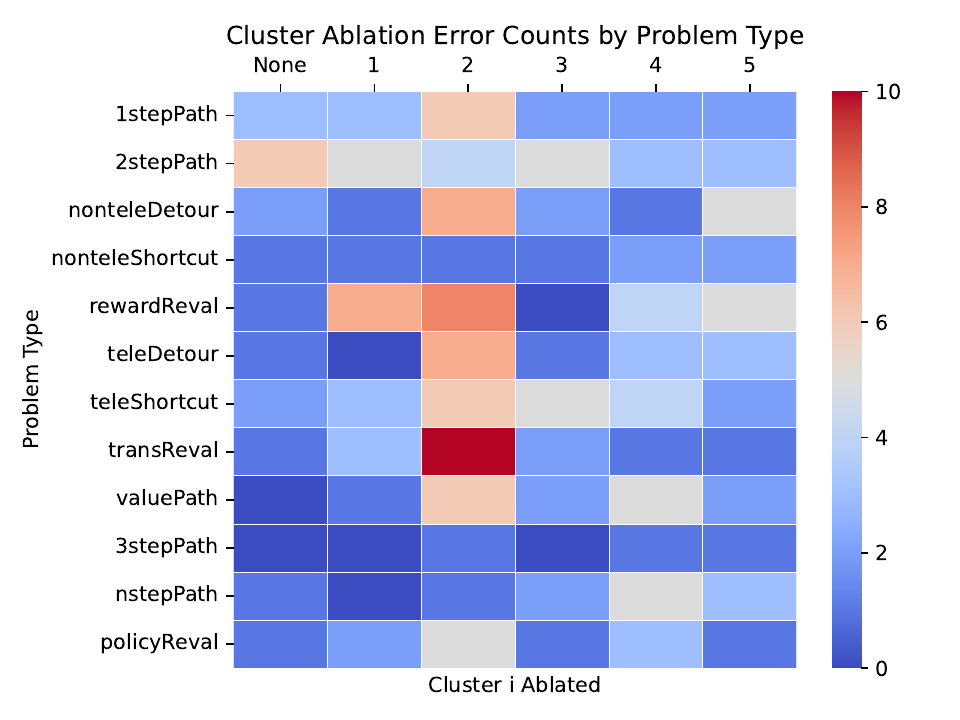}\label{fig:ablation_error_count}}
 \caption{Ablating ICL examples by failure mode. We investigated which examples, from which failure mode cluster defined in Section ~\ref{sec:fail_mode_cat}, were crucial to the success of ICL prompts. To do so, we used ablation to probe the effect of each cluster of examples on the effectiveness of the ICL prompt. That is, we systematically kept everything in a given ICL prompt the same but removed examples that belonged to a specific cluster of failure modes. We then re-run the evaluator with the ablated ICL. (a) Ablation of examples from each cluster of failure mode led to dissimilar performance compared with performance using the original ICL (pairwise comparison, Cohen's Kappa ~\cite{cohen_kappa}). (b) Demonstrated the influence of ablated examples from each failure mode cluster on the Evaluator's behavior. (c) Error counts by problem type for each cluster ablation. The original ICL prompt yielded a total of 19 errors. Ablating Cluster 2 had the largest impact on performance, yielding 62 errors, while ablating Cluster 3 had the least impact on performance, yielding 23 incorrect predictions.}\label{fig:ablation_hmaps}
\end{figure}

Each of the above categories include several ICL examples in~\cite{cogeval}, and we can form five ICL clusters accordingly. To better visualize and assess pairwise similarity of these ICL clusters we leveraged SBERT sentence transformer~\cite{sentence_bert}, a state-of-the-art model for text semantic similarity analysis. Each ICL example is encoded into a $384$-dimensional embedding, and is projected into a $2$D tSNE map as shown in Figure~\ref{fig:tsne}. The distinction between the clusters is noticeable, and the clusters that are contextually closer are adjacent, e.g., Cluster 1 (Partial Correctness) and Cluster 2 (Use of Keywords) or Cluster 4 (Gibberish Response) and Cluster 5 (Parroting Response).  

To identify the most relevant ICL clusters, in what follows, we conduct a systematic cluster ablation test where in each step we remove a cluster from the set of ICL examples and analyze the change in the evaluator performance. The test results are summarized in Figure~\ref{fig:ablation_test}. It is evident that the ablation of Cluster 2 (ICLs that include Keywords) from the set of ICL examples has the most significant impact on the performance of the evaluator. Namely, robustifying the evaluator using ICL against its 2nd failure mode (i.e., Deceived by Keywords) appears to be most crucial in CogEval.

To further investigate the similarities (and dissimilarities) between the ICL clusters we focused deeper on the score labels generated by the evaluator when a cluster is ablated.

Figure ~\ref{fig:ablation_error_count} demonstrates that the benefits of ICL are non-additive. This can be observed by the model performance on teleDetour where ablating cluster 1 increases the performance but ablating cluster 2 decreases the performance. Further we see an asymmetry to the importance per ICL cluster. Ablating Cluster 2 had a substantial impact increasing the total number of errors by 226 percent while ablating Cluster 3 increased errors by 21 percent. This highlighted that relationships between the ICL examples also changed the evaluator’s behavior. Therefore, it is important when picking ICL examples to not only focus on the relationship between the example and the question but also potential relationships which may exist between ICL examples.


\section{Experiment 2: Auditing Text Summarization}
\begin{figure}[!t]
    \centering    \includegraphics[width=0.65\textwidth]{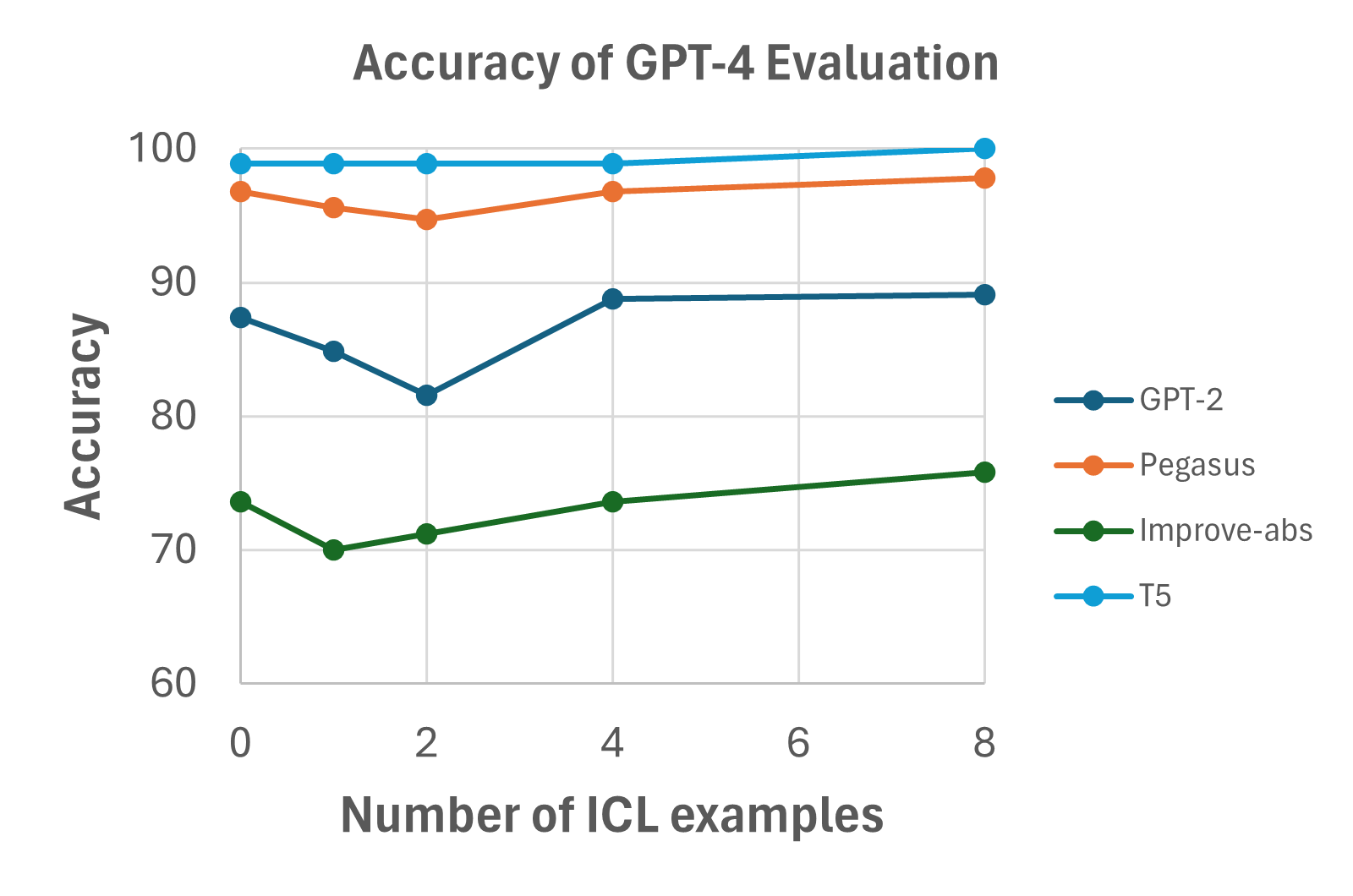}
    \caption{Changes in GPT-4's evaluation accuracy with increasing number of ICL examples. GPT-4 evaluated the factual correctness of news summarization (also called "consistency" in the original database) by four LLMs: GPT-2, T5, Pegasus and Improve-abs. We compared each LLM's generated summaries with human labels to compute factual correctness. We observe that adding 4 or more ICL examples improves GPT-4's evaluation of summarization by all four LLMs.}
    \label{fig:acc_summ}
\end{figure}

In this section, we evaluate the factual correctness of the summarization. We use the SummEval data \cite{fabbri2020summeval}, 
which is based on the CNN/DailyMail dataset \cite{hermann2015teaching}. It contains human ratings on four aspects of each summary: \texttt{fluency},
\texttt{coherence}, \texttt{consistency} and \texttt{relevance} on a scale of $1-5$. We focus on the factual correctness of the summary, given by the \texttt{consistency} score. We consider a summary to be factually incorrect if the \texttt{consistency} score is less than $2.5$, otherwise it is factually correct. We select the summaries generated by GPT-2, T5, Pegasus and Improve-abs\footnote{\url{https://github.com/Yale-LILY/SummEval}}. Given the document and its summary, we use GPT-4 for evaluation. It outputs the score as $1$ if the summary if factually correct, otherwise the score is $0$ for summaries with hallucinations. Similar to G-Eval \cite{liu2023gpteval}, which uses LLM like GPT-4 and  GPT-3.5 (text-davinci-003) to score the documents and summaries, and measures the correlations with human score, we have the following chain-of-thoughts (CoT) prompt for evaluation:
\begin{tcolorbox}[width=\textwidth]
\emph{Evaluation Criteria:
Consistency (0 or 1) - the factual alignment between the summary and the summarized source. A factually consistent summary contains only statements that are entailed by the source document. Annotators assign score as 1 if the summary is factually correct else the score is 0.
Evaluation Steps:\\
1. Read the news article carefully and identify the main facts and details it presents.\\
2. Read the examples of document, summary, and the scores for factual correctness.\\
3. Read the summary and compare it to the article. Check if the summary contains any factual errors that are not supported by the article.\\
4. Assign a score for consistency based on the Evaluation Criteria.}
\end{tcolorbox}
The examples consist of the source text, summary and consistency score as shown below:
\begin{tcolorbox}[width=\textwidth]
\emph{"Source text": "Paul Merson has restarted his row with Andros Townsend after the Tottenham midfielder was brought on with only seven minutes remaining in his team 's 0-0 draw with Burnley on Sunday..."\\
"Summary": "Paul merson was brought on with only seven minutes remaining in his team 's 0-0 draw with burnley...."\\
"Consistency": 0\\
\\
"Source text": "Nathan Hughes on Friday night had his ban for accidentally knocking out George North sensationally over-turned on appeal...."\\
"Summary": "Nathan hughes had his ban for accidentally knocking out george north..."\\
"Consistency": 1
}
\end{tcolorbox}
We use GPT-4 to score the consistency of a given source text and summary and then we measure the accuracy of these predictions against the human annotations. 
Figure~\ref{fig:acc_summ} shows the accuracy of GPT-4 predictions for factual correctness with the number of ICL examples in the prompt for SummEval dataset where the accuracy is measured with respect to human labels for factual correctness (also called the \texttt{consistency} aspect in the database). We found that with 4 or more demonstration examples, the accuracy does improve for the evaluator.

\section{Discussion and future directions}

LLMs are widely used to automate evaluation of educational and medical text, in spite of displaying failure modes that pose ethical and technical challenges \cite{Yan2023edu}. Thus, despite the extensive utility of LLMs in tasks such as grading papers, assessing summarized medical documents, or analyzing user responses, their text evaluation capabilities still require a thorough audit and improvement. To address this challenge, we propose ALLURE, a systematic protocol for auditing and improving LLM-based evaluation of text using iterative ICL. 

ALLURE compares LLM-generated evaluations with annotated data, iteratively incorporating instances of significant deviation into the evaluator's ICL prompts (context). The approach leverages in-context learning (ICL) to enhance and improve the robust evaluation of text by LLMs. We improve and investigate improvements due to ICL in two ways. First, ALLURE iteratively adds corrected examples of failure modes to the ICL prompts in order to improve evaluation. Second, we use ablation to understand which specific examples (related to the same cluster of failure modes) were crucial to the improvements. Through this iterative process and ablation, our goal was to refine the performance of the evaluator LLM with minimal ICL, ultimately reducing the reliance on human annotators in the evaluation process. 

We found that while ICL examples can improve GPT-4's evaluation performance, merely increasing the number of examples does not improve performance in an additive manner (both at a problem class and overall). In fact, increasing the number of examples in ICL initially leads to a dip in ICL-based improvement, but improves after a certain amount is added (decrease up to 30 and then improvement overall at 45 number of ICL). Furthermore, we find that different sets or clusters of "similar" ICL examples (categorized based on different failure modes) yield asymmetric improvements in GPT-4's evaluation accuracy. This points to the importance of analyzing selected examples for ICL prompts holistically, considering the relationship of different clusters of failure modes and categories of examples. 

Future directions include a thorough analysis of attention mechanisms underlying the improvements due to ICL. Another direction is a thorough analysis of the changes induced by ALLURE's ICL on the underlying embeddings, or latent representations. It would be interesting to come up with metrics to measure the contributions of ICL to changes in attention and embeddings that mediate the improvements. Finally, while ALLURE reduces reliance on humans by using a systematic protocol, one limitation of the current approach is that the ICL prompts still require a human in the loop. An important future direction would be to automate the entire achieve out of distribution (OOD) generalization. One possible approach would be to automate extraction of the relevant features and examples of failure modes, automate the iterative process of ICL generation and re-running of the evaluation, and automate the auditing of GPT-4's evaluation.

To summarize, we propose ALLURE, a systematic protocol for Auditing and Improving Large Language Model based evaluation of text. We expect that ALLURE will benefit diverse applications of LLMs in various domains related to the evaluation of textual data, potentially enhancing productivity in these fields. By addressing the limitations of LLMs and improving their text evaluation capabilities, ALLURE represents a significant step forward in the optimization and adaptation of large language models for a wide range of evaluation tasks.

\begin{acks}
We would like to thank Richard Ciapala for engineering support.
\end{acks}

\clearpage
\bibliographystyle{ACM-Reference-Format}
\bibliography{MAIN_BIB}


\clearpage
\appendix

\section{Appendix}

\subsection{ICM Examples}

In Experiment 1, the cognitive maps are inspired by a set of tasks adapted from existing human behavioral experiments \cite{Schapiro2013-mx, Momennejad2017-wr, Momennejad2018-zd, Momennejad2019-tf, Pudhiyidath2022-sw}. Navigating cognitive maps requires adaptive multi-step planning using compressed representations of the environment. In the following we present the distribution of the ICL examples over the required planning skills, described in Table~1 in~\cite{cogeval}.

\begin{figure}[H]
  \centering \includegraphics[width=0.5\textwidth]{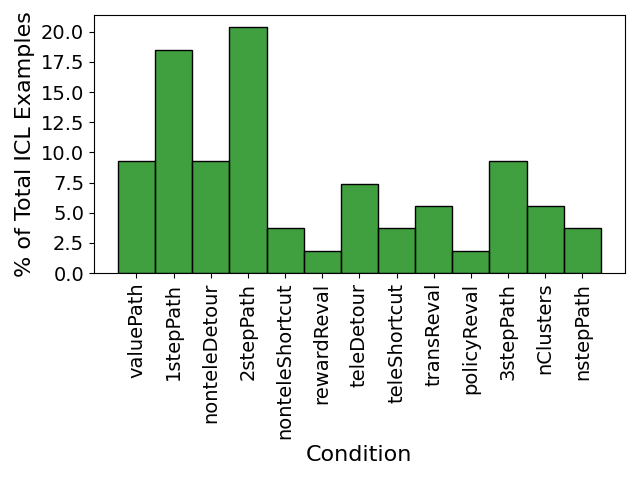}
  \caption{Distribution of ICL examples over the planning skills described in Table~1 in~\cite{cogeval}. Interestingly, "path" planning skill sets were the dominant failure modes, which resulted in more ICL examples for those skill sets.} \label{fig:ICL_conditions}
\end{figure}
\clearpage

\subsection{tSNE Analysis}
\renewcommand{\scale}{0.4}
\begin{figure}[H]
	\centering
    \subfloat[][]{\includegraphics[width=\scale\linewidth]{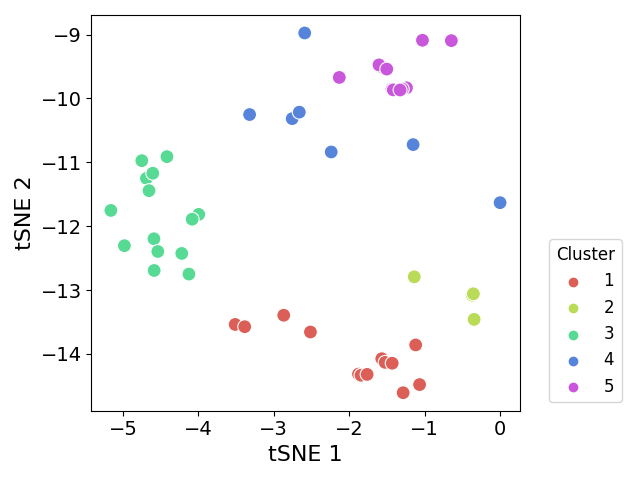}}
	\qquad
	\subfloat[][]{\includegraphics[width=\scale\linewidth]{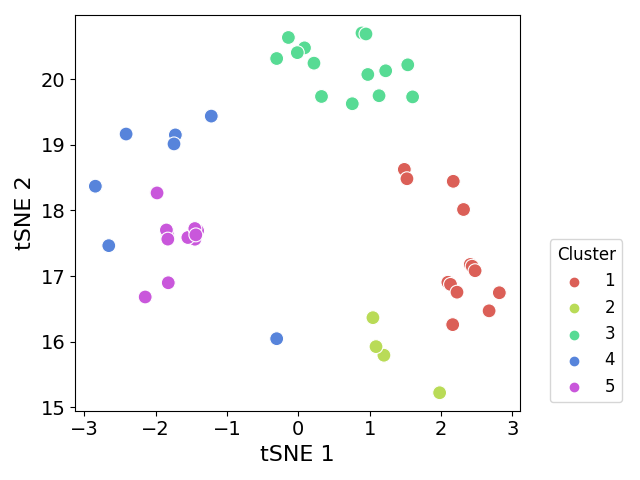}}
    \qquad
	\subfloat[][]{\includegraphics[width=\scale\linewidth]{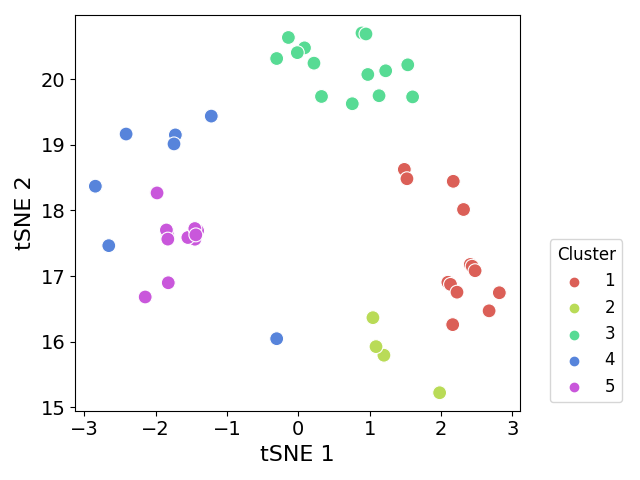}}
    \qquad
	\subfloat[][]{\includegraphics[width=\scale\linewidth]{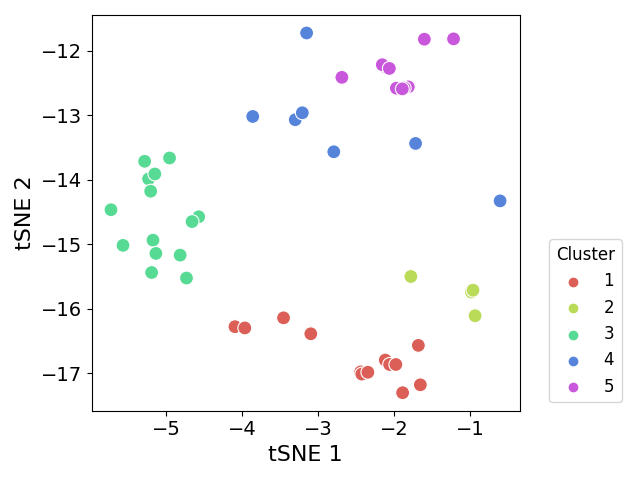}}
 \caption{Several runs of tSNE over the ICL example embeddings. Given the stochastic nature of tSNE, the clusters are still distinguishable and similar ICL examples are adjacent to each other.}\label{fig:tSNE_5runs}
\end{figure}

\clearpage


\end{document}